\documentclass[journal]{IEEEtran}
\usepackage{graphicx}
\usepackage{amsmath}
\usepackage{booktabs}
\usepackage{multirow}
\usepackage{placeins} 

\ifCLASSINFOpdf
\else
\fi
\begin{document}
%
\title{Sparse Multi-Modal Transformer with Masking for Alzheimer’s Disease Classification}

\author{Cheng-Han~Lu \and Pei-Hsuan~Tsai}
\maketitle

\begin{abstract}
Transformer-based multi-modal intelligent systems often suffer from high computational and energy costs due to dense self-attention, limiting their scalability under resource constraints. This paper presents SMMT, a sparse multi-modal transformer architecture designed from a system-level perspective to improve efficiency and robustness. Building upon a cascaded multi-modal transformer framework, SMMT introduces cluster-based sparse attention to reduce computational complexity to $\mathcal{O}(n \log n)$ and modality-wise masking to enhance robustness against incomplete inputs. The architecture is evaluated using Alzheimer’s Disease classification on the ADNI dataset as a representative multi-modal case study. Experimental results show that SMMT maintains competitive predictive performance while significantly reducing training time, memory usage, and energy consumption compared to dense-attention baselines, demonstrating its suitability as a resource-aware architectural component for scalable intelligent systems.
\end{abstract}

\begin{IEEEkeywords}
Alzheimer’s Disease, Multi-modal Fusion, Sparse Attention, Transformer Architecture, Masked Learning, Human Digital Twin, Medical Representation Learning, Clinical Data Integration, Low-Resource Generalization, Deep Neural Networks.
\end{IEEEkeywords}

%
\IEEEpeerreviewmaketitle

\section{Introduction}
%
%
%
%
\IEEEPARstart{A}{lzheimer’s} Disease (AD) is a progressive neurodegenerative disorder that leads to severe cognitive decline and significantly impairs patients’ quality of life. Timely and accurate diagnosis is essential for initiating early interventions that may delay disease progression and improve clinical outcomes. However, this remains a complex challenge due to the subtle nature of early-stage symptoms and significant inter-patient variability [1]–[4]. In response, recent advances in artificial intelligence (AI) and deep learning have enabled models to leverage multi-modal information—such as structural MRI, cognitive scores, and genetic markers—for supporting AD diagnosis.

Despite this progress, two critical challenges persist in AI-based AD classification. First, the size of high-quality medical datasets is often limited, particularly for early-stage AD and Mild Cognitive Impairment (MCI), due to high acquisition costs and patient privacy constraints [5]–[7]. Second, clinical data is inherently heterogeneous and frequently incomplete, with missing modalities being common in real-world clinical settings. To mitigate these limitations, prior research has explored various strategies, including data augmentation [5], robust feature representation learning [11], modality-invariant modeling, and imputation techniques to compensate for missing data [6], [14]. However, these approaches are typically constrained by scalability and may struggle to maintain performance when applied to complex multi-modal inputs.

Among the most promising solutions in recent years is the Cascaded Multi-Modal Mixing Transformer (3MT) [8], which represents a state-of-the-art approach for AD diagnosis using hybrid fusion. By combining modality-specific Transformer encoders with cascaded cross-attention layers, 3MT effectively captures both intra- and inter-modal relationships, enabling rich and dynamic feature representation across heterogeneous data sources. Compared to conventional CNN-based models and early/late fusion schemes, 3MT has shown superior classification performance, especially when dealing with structured and unstructured medical data.

However, despite its effectiveness, 3MT still suffers from key limitations that hinder its broader applicability. First, it relies heavily on dense self-attention mechanisms, resulting in computational complexity that scales quadratically with the sequence length $\mathcal{O}(n^2)$. This poses challenges for scaling the model to large inputs or resource-constrained settings. Second, 3MT lacks explicit mechanisms to address missing modalities, which frequently occur in clinical practice. As a result, its performance tends to degrade in small-sample or incomplete-data scenarios, limiting its generalizability in real-world healthcare environments.

To address these limitations, we propose the Sparse Multi-Modal Transformer with Masking (SMMT) a hybrid fusion framework designed for efficient and robust AD classification, particularly in low-resource settings. SMMT introduces two key innovations: (1) a cluster-based sparse attention mechanism that significantly reduces computational cost by limiting attention computation within token clusters, and (2) a modality-wise masking strategy that improves generalization by simulating missing modalities during training. Through extensive experiments on the ADNI dataset, we demonstrate that SMMT achieves superior diagnostic performance, reaching 97.05\% accuracy on full data and 84.96\% accuracy using only 20\% of the data. Additionally, SMMT reduces training energy consumption by 40.4\% compared to 3MT, highlighting its potential for scalable and sustainable deployment in real-world clinical environments.

\section{Related Work}
AI has become an essential tool in the diagnosis and prognosis of AD, with deep learning methods showing promising results across various data modalities [3]. Early approaches predominantly relied on single-modality data, such as structural MRI, to classify AD and cognitively normal (CN) subjects using convolutional neural networks (CNNs) or support vector machines [22]. However, such models are limited in their ability to capture the full complexity of AD progression, which often requires a more holistic integration of imaging, cognitive, and genetic information.

In response to the limitations of single-modality approaches, multi-modal learning has emerged as a powerful strategy, enabling models to integrate complementary features from heterogeneous data sources. Recent studies have demonstrated the effectiveness of combining MRI with clinical scores (e.g., MMSE, CDR) and demographic or genetic information (e.g., APOE genotype) for more accurate and robust prediction. For example, Zhang et al. [19] proposed a multi-modal deep learning framework to jointly learn from neuroimaging and clinical assessments, while Suk et al. [15] used stacked autoencoders to extract shared representations across modalities. Similarly, Golovanevsky et al. [18] applied attention-based fusion to capture informative relationships across modalities for Alzheimer's diagnosis.

Despite these advances, several key challenges remain. Real-world clinical datasets are often limited in size and contain missing modalities due to inconsistent acquisition protocols, high imaging costs, or privacy concerns. To address these issues, researchers have developed three main fusion strategies for multi-modal learning: early fusion, late fusion, and hybrid (intermediate) fusion.

Early fusion combines raw or low-level features from multiple modalities at the input level, allowing the model to learn joint representations from the beginning. For example, Zhang and Shi [12] proposed a model that concatenates multi-modal neuroimaging features before feeding them into a deep learning classifier, achieving competitive performance when all modalities are present. However, this approach is highly sensitive to misaligned or missing data, which often degrades performance in real-world applications.

Late fusion processes each modality independently through dedicated subnetworks and merges their outputs at the decision level, such as by averaging predictions or using a meta-classifier. Dwivedi et al. [16] developed a deep learning framework that separately processes MRI and clinical data and fuses the outputs using a meta-classification layer. Similarly, Zuo et al. [17] introduced an adversarial hypergraph fusion model that maintains modality-specific representations while combining predictions in a robust manner. These late fusion methods offer flexibility, especially under partial modality availability, but often overlook cross-modal interactions that could enrich the learned representations.

Hybrid (intermediate) fusion has emerged as a more effective approach for multi-modal integration by combining modalities at intermediate stages of the network. This strategy preserves each modality’s unique structural properties while enabling the model to learn rich inter-modal relationships. Among hybrid fusion architectures, Transformer-based models have gained increasing attention due to their ability to model long-range dependencies through self- and cross-attention mechanisms. Notable examples include Triformer [13], which processes each modality with a dedicated Transformer encoder followed by cross-attention fusion; and the models proposed by Lee et al. [14] and Aghili et al. [15], which further enhance robustness by incorporating cognitive and demographic data or imputation-aware attention for handling missing modalities. Among these efforts, the 3MT model [8] stands out as a representative approach: it encodes each modality using modality-specific Transformers and fuses them through a cascaded sequence of cross-attention layers. 3MT has demonstrated strong performance on AD classification tasks when complete multi-modal inputs are available, serving as a solid foundation for further enhancements in both efficiency and robustness.

While 3MT demonstrates strong performance under complete multi-modal conditions, its reliance on dense self-attention and the absence of explicit mechanisms for handling missing modalities limit its applicability in real-world clinical settings. These observations highlight the need for more efficient and robust fusion architecture that can better accommodate incomplete data and scale to larger input spaces. Motivated by these gaps, the following section introduces our proposed model, which addresses these limitations through architectural innovations in attention and modality handling.

\section{Methodology}
The method extends the baseline 3MT architecture with two key enhancements: sparse attention and masking-based regularization.

\subsection{Baseline Architecture: Cascaded Multi-Modal Mixing Transformer (3MT)}
The baseline of our framework is the 3MT architecture [8], designed for multi-modal AD classification tasks. It processes heterogeneous data types—such as MRI images, cognitive scores, and genetic markers—by encoding each modality separately and then fusing them through a cascaded attention mechanism.

\begin{figure*}[!t]
    \centering
    \includegraphics[width=\linewidth]{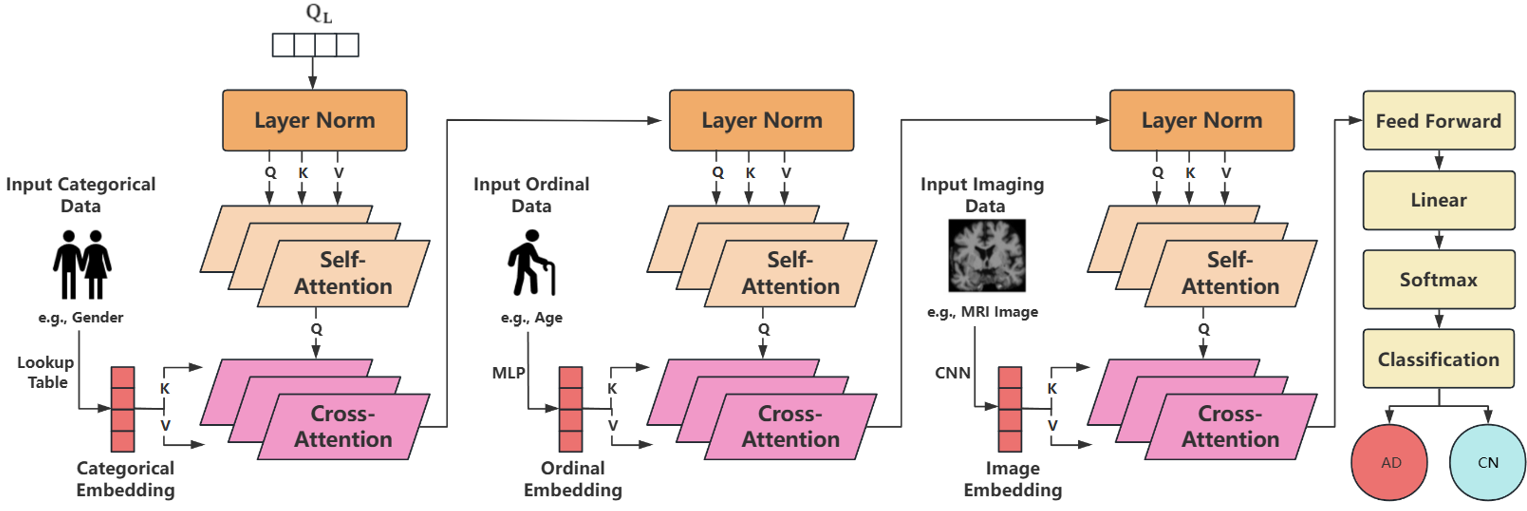}
    \caption{Architecture of the baseline 3MT. Each modality is encoded into a 512-dimensional feature space using modality-specific encoders. Self-attention captures intra-modal dependencies, and cross-attention allows inter-modal fusion. The final representation is passed to a classifier for AD/CN prediction.}
    \label{fig:3mt_architecture}
\end{figure*}

As illustrated in Fig.~\ref{fig:3mt_architecture}, each modality is projected into a common 512-dimensional latent space using modality-specific encoders:

\begin{itemize}
    \item maging features are extracted using a CNN with four convolutional blocks followed by a fully connected layer.
    \item Numerical clinical scores (e.g., MMSE, CDR, FAQ, age) are transformed using a multi-layer perceptron (MLP) composed of three fully connected layers with ReLU activations.
    \item Categorical data (e.g., APOE genotype, sex) are mapped to 32-dimensional embeddings via a learnable lookup table, concatenated, and linearly projected to match the shared feature space.
\end{itemize}

Within the Transformer pipeline, each modality undergoes intra-modal \textit{self-attention} to capture internal relationships. \textit{Cross-attention} then integrates inter-modal dependencies by injecting feature representations from one modality into another. This cascaded design enables flexible and dynamic fusion, allowing for comprehensive modeling of complementary data sources.

Formally, the self-attention operation in each Transformer layer is defined as:

\begin{equation}
\mathrm{Attention}(Q, K, V) =
\mathrm{softmax} \left( \frac{QK^\top}{\sqrt{d_k}} \right) V
\end{equation}

where $Q$, $K$, $V \in \mathbf{R}^{n \times d}$ denote the query, key, and value matrices respectively, and $d_k$ is the dimension of each head. For cross-attention, the query originates from one modality while key and value originate from another. The 3MT applies this architecture in a cascaded sequence to iteratively enrich the learned representation.

Cross-attention for inter-modal fusion is defined similarly, but $Q$ comes from one modality and $K$, $V$ come from another:

\begin{equation}
\mathrm{CrossAttention}(Q, K, V) =
\mathrm{softmax} \left( \frac{QK^\top}{\sqrt{d_k}} \right) V
\end{equation}

\begin{figure*}[t]
    \centering
    \includegraphics[width=\textwidth]{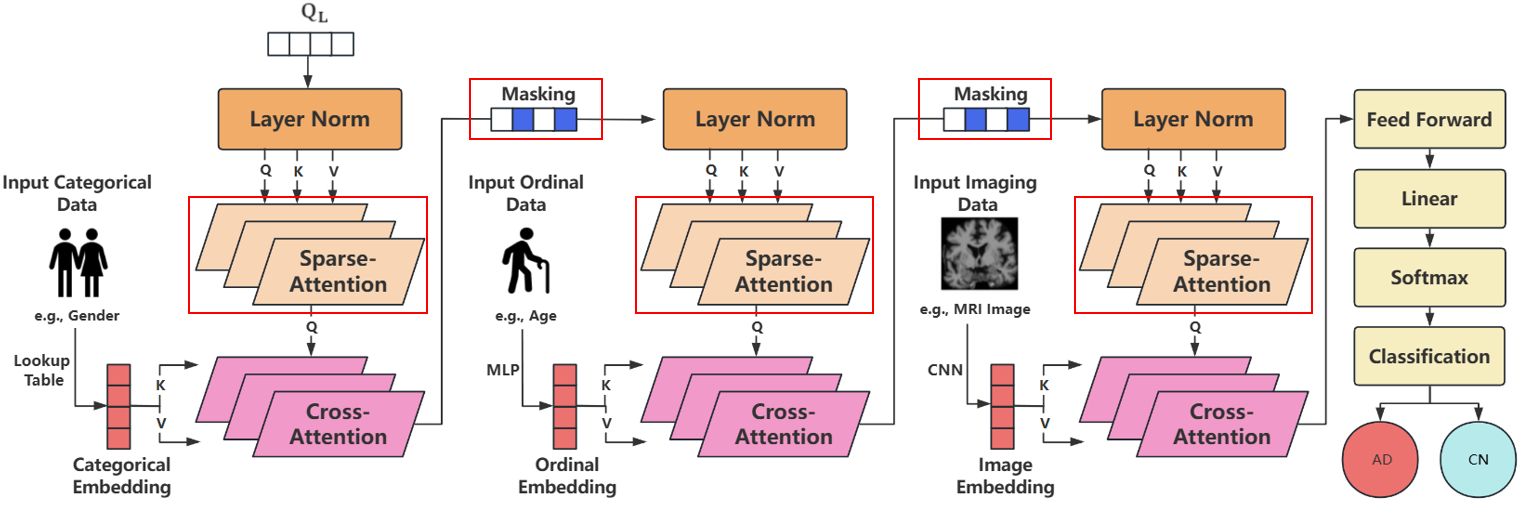} 
    \caption{Proposed architecture integrating Sparse Attention and Masking. Compared to the original 3MT model shown in Fig.~1, we introduce two key enhancements, both highlighted by red rectangles: (1) Sparse Attention replaces traditional dense self-attention to reduce computational complexity while preserving performance, and (2) Masking modules are inserted between modalities to randomly drop modality-specific features during training, improving generalization under missing or limited data conditions. These enhancements are applied across all modalities (categorical, ordinal, and imaging), and the final fused features are passed through a feed-forward classification head to predict AD vs. CN.}
    \label{fig:smmt_architecture}
\end{figure*}

After attention-based fusion, the resulting modality-specific representations are aggregated and passed to a Multi-Layer Perceptron (MLP) classifier for binary classification (AD vs. CN). The overall training objective is to minimize cross-entropy loss between the predicted and ground truth labels. Backward propagation is applied end-to-end after all modalities are processed and fused, allowing gradients to update all components, including CNN encoders, MLPs, attention layers, and embedding tables. This unified optimization strategy ensures effective learning from heterogeneous data and enhances generalization performance across real-world clinical conditions.

\subsection{Proposed Model: SMMT (Sparse Multi-Modal Transformer with Masking)}
In response to the shortcomings of 3MT, we develop the SMMT, which integrates two principal enhancements:
\begin{itemize}
    \item Sparse Attention: Reduces computational complexity from $\mathcal{O}(n^2)$ to $\mathcal{O}(n \log n)$ by restricting attention to tokens within the same cluster. Clustering is performed via K-Means using query vectors with $k = \log_2 n$ clusters.
    \item Masked Representation Learning: Applies random masking to fused features during training to enhance generalization and robustness.
\end{itemize}

While K-Means clustering introduces an additional computational step, we choose it over heuristic or rule-based patterns due to its adaptive and data-driven nature. In multi-modal learning scenarios, fixed attention spans or proximity-based heuristics may fail to capture meaningful semantic relationships between modalities such as image patches and structured tabular features. K-Means offers a principled approach to token grouping based on learned query vector similarities, resulting in semantically aligned clusters for efficient attention computation.

From a complexity standpoint, the K-Means step incurs a cost of \( O(nkdi) \), where \(n\) is the token length, \(k = \log_2 n\), \(d\) the feature dimension, and \(i\) the number of iterations. In our implementation, clustering is performed once per batch on the GPU and reused for all attention heads, resulting in minimal overhead relative to the savings from reduced attention computations. Runtime profiling further confirms that the overall attention module remains faster than full self-attention, validating our choice as an effective balance between flexibility and efficiency.

As illustrated in Fig. 2, our architecture retains the cascaded design of the 3MT pipeline, which performs modality-specific processing followed by inter-modal feature fusion. It replaces the intra-modal dense self-attention with cluster-based sparse attention and applies masking to the output of each modality after cross-attention to improve robustness under small-sample conditions.

\subsubsection*{Sparse Attention Formulation}

\begin{equation}
\begin{aligned}
&\mathrm{SparseAttention}(Q, K, V) = \sum_{j \in \mathcal{C}(i)} \alpha_{ij} \, V_j , \\
&\alpha_{ij} = \frac{\exp(Q_i \, K_j^\top)}{\sum_{j' \in \mathcal{C}(i)} \exp(Q_i \, K_{j'}^\top)}
\end{aligned}
\end{equation}

Here, $Q_i \in \mathbf{R}^{1 \times d}$ is the query vector corresponding to token $i$, and $K_j$, $V_j \in \mathbf{R}^{1 \times d}$ are the key and value vectors for token $j$. $\mathcal{C}(i)$ denotes the set of tokens that belong to the same cluster as token $i$, determined via K-Means clustering over query embeddings.

The attention weight $\alpha_{ij}$ quantifies the influence of token $j$ on token $i$ within its cluster. The denominator uses a separate index $j'$ to sum over all tokens in the same cluster considered in the normalization step, ensuring attention scores are normalized locally. Thus, $j$ is the current token being attended to, and $j'$ represents all tokens within $\mathcal{C}(i)$.

\subsubsection*{Masked Representation Learning}

\begin{equation}
\tilde{x} = x \odot m, \quad m \sim \mathrm{Bernoulli}(1 - r)
\end{equation}

Here, $r$ is the masking ratio (e.g., 0.3), and $m$ is a binary mask sampled independently for each dimension from a Bernoulli distribution with success probability $1 - r$. The operator $\odot$ denotes element-wise multiplication, which zeroes out masked dimensions of the input feature vector. This simulates partial feature corruption during training and helps improve generalization under data-scarce conditions.

This design ensures both computational efficiency and robustness to limited data. Sparse attention reduces memory and compute overhead, while masked representation learning simulates modality corruption during training, allowing the model to develop resilience and better generalize to unseen or missing data.

\section{Experiments and Results}
\subsection{Experimental Setup}
All experiments were implemented in PyTorch and executed on a local workstation equipped with an NVIDIA GeForce RTX 3060 GPU (12GB memory). The key training configurations and model hyperparameters are summarized in Table~\ref{tab:setup}.

\begin{table}[htbp]
\centering
\caption{Experimental Setup and Hyperparameters}
\label{tab:setup}
\begin{tabular}{ll}
\toprule
\textbf{Category} & \textbf{Setting} \\
\midrule
Framework & PyTorch \\
Hardware & NVIDIA RTX 3060 GPU (12GB) \\
Epochs & 50 \\
Batch Size & 8 \\
Optimizer & Adam \\
Learning Rate & $1 \times 10^{-3}$ \\
Loss Function & Cross Entropy \\
Latent Space Dim. & 512 \\
Q/K/V Dimension & 512 (shared across modalities) \\
Masking Strategy & Modality-level random masking ($r = 0.3$) \\
Repeatability & 5 runs with different seeds \\
\bottomrule
\end{tabular}
\end{table}

\subsection{Data Preprocessing}
To prepare the data for model training and evaluation, we extracted T1-weighted 2D MRI slices and aligned them with structured clinical metadata from the ADNI-1 and ADNI-2 cohorts [23]. Only subjects with a confirmed diagnosis of either AD or CN were retained. Subjects with Mild Cognitive Impairment (MCI) or inconsistent labels were excluded. In total, 12{,}680 MRI slices were collected for the final dataset.

All MRI images were registered to the MNI152 template using linear alignment, then resized to $256 \times 256$ pixels, normalized to the $[0, 1]$ intensity range, and converted to RGB format to meet the input requirements of convolutional neural networks. Only MPRAGE T1-weighted scans were retained, and duplicates or ambiguous records were removed based on the modality and scan description.

Structured clinical features included MMSE Total Score, Global CDR, FAQ Total Score, Age, Sex, and APOE genotype. These features were selected for their clinical relevance and demonstrated effectiveness in previous AD diagnostic studies [9], [10]. Clinical entries were matched with MRI scans via the Image ID field and synchronized based on the nearest Study Date.

\begin{table}[htbp]
\centering
\caption{Demographic Information of Study Subjects of Three Datasets for the AD Classification Task}
\label{tab:demographics}
\begin{tabular}{lllll}
\toprule
\textbf{Category} & \textbf{Dataset} & \textbf{No. of Subjects} & \textbf{Age Range} & \textbf{Female/Male} \\
\midrule
\multirow{2}{*}{CN} & ADNI-1 & 1{,}634 & 62.7--92.8 & 806 / 828 \\
                   & ADNI-2 & 395    & 76.6--95.3 & 165 / 230 \\
\midrule
\multirow{2}{*}{AD} & ADNI-1 & 858    & 55.7--93.0 & 394 / 464 \\
                   & ADNI-2 & 104    & 55.9--88.5 & 45 / 59 \\
\bottomrule
\end{tabular}
\end{table}

\subsection{Evaluation Metrics}

To comprehensively evaluate the effectiveness of our model, we employ two categories of metrics: classification performance and computational sustainability. The former quantifies the diagnostic accuracy of the model, while the latter assesses its environmental and computational efficiency during training.

\subsubsection{Classification Metrics}
We adopt standard evaluation metrics including Accuracy, Precision, Recall (Sensitivity), and F1-score to assess model performance on binary classification between AD and CN subjects. These metrics are defined as:

\begin{equation}
\mathrm{Precision} = \frac{TP}{TP + FP}
\end{equation}

\begin{equation}
\mathrm{Recall} = \frac{TP}{TP + FN}
\end{equation}

\begin{equation}
F_1\text{-score} = \frac{2 \times \mathrm{Precision} \times \mathrm{Recall}}{\mathrm{Precision} + \mathrm{Recall}}
\end{equation}

\begin{equation}
\mathrm{Accuracy} = \frac{TP + TN}{TP + TN + FP + FN}
\end{equation}

where $TP$, $TN$, $FP$, and $FN$ represent true positives, true negatives, false positives, and false negatives, respectively. These metrics provide a more complete picture of model behavior, particularly under class imbalance, which is common in medical datasets.

\subsubsection{Computational Sustainability Metrics}
To assess the environmental impact and training efficiency of each model, we estimate total energy consumption and carbon emissions using the \texttt{CodeCarbon} framework. \texttt{CodeCarbon} is an open-source tool that monitors hardware usage and computes emissions based on the carbon intensity of the power grid [24].

\texttt{CodeCarbon} estimates the total carbon dioxide ($CO_2$) emissions as follows:

\begin{equation}
CO_2\ \mathrm{emissions\ (kg)} = E_{\mathrm{consumed}} \times CI
\end{equation}

Where $E_{\mathrm{consumed}}$ represents the total energy consumption in kilowatt-hours ($kWh$), measured during training based on CPU and GPU usage, and $CI$ denotes the carbon intensity of the power grid ($kgCO_2/kWh$), based on regional defaults or real-time tracking.

\texttt{CodeCarbon} estimates $E_{\mathrm{consumed}}$ by monitoring:
\begin{itemize}
    \item CPU power draw: calculated from processor power usage per core and utilization over time.
    \item GPU power draw: obtained through the NVIDIA Management Library (NVML), which directly reads GPU power usage in watts during model training.
    \item RAM energy consumption: estimated using empirical wattage per GB and memory utilization.
\end{itemize}

The tool logs real-time usage during training and aggregates the consumption from all active components. This allows for model-wise comparison of both training efficiency (measured in minutes) and energy cost (measured in kWh and corresponding ($CO_2$)  emissions), providing a comprehensive view of each model’s computational footprint.

\subsection{Classification Results}
To comprehensively evaluate the effectiveness of our model, we compare SMMT with several representative and SOTA approaches. The 3MT baseline [8] serves as the primary reference model, known for its cascaded attention design and strong performance in multi-modal AD classification. ADDFformer [20] is a Transformer-based method tailored for structural MRI, achieving competitive accuracy in imaging-only AD detection. We also include a general-purpose multi-modal fusion network [21] that employs joint feature learning without attention mechanisms, and CNN-only models [22] that rely exclusively on MRI, serving as a single-modality baseline. These comparison models span a spectrum of design paradigms from imaging-only to attention-based fusion, providing a well-rounded benchmark for assessing the efficiency and robustness of our proposed architecture.

\begin{figure}[htbp]
    \centering
    \includegraphics[width=\linewidth]{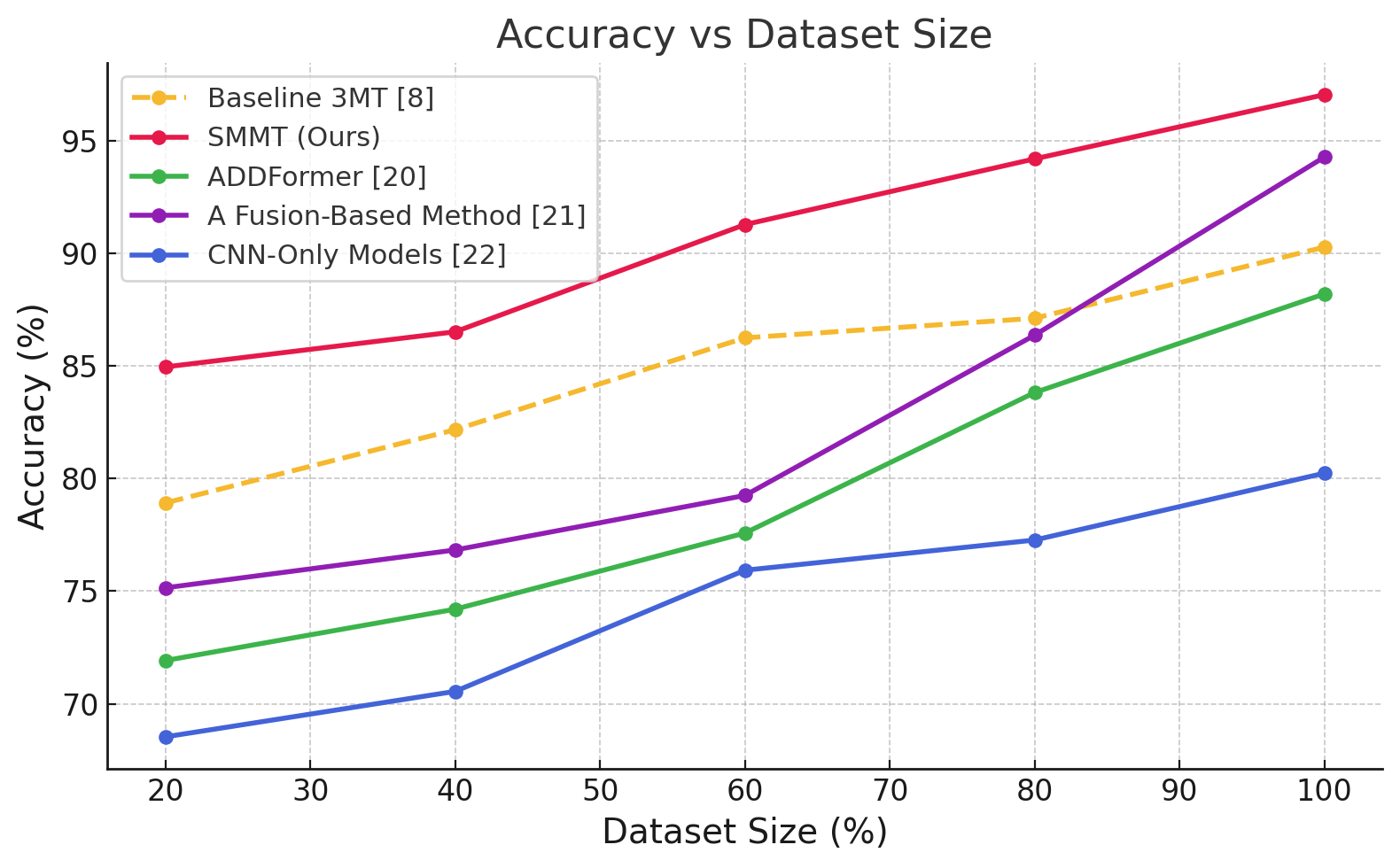} 
    \caption{Accuracy (\%) vs. dataset size for SMMT and baseline models. Our method consistently outperforms existing approaches, particularly in low-data regimes.}
    \label{fig:accuracy_curve}
\end{figure}

Fig.~\ref{fig:accuracy_curve} and Table~\ref{tab:acc_results} illustrate the classification accuracy at different dataset sizes. Our method consistently outperforms all baselines, especially under low-data conditions. At 20\% data, SMMT achieves 84.96\% accuracy, while the best-performing baseline model reaches only 78.92\%. As the dataset size increases, the accuracy advantage of our model remains significant. This highlights the robustness and data-efficiency of our proposed sparse attention and masking mechanisms.

\begin{table}[htbp]
\centering
\caption{Accuracy (\%) Comparison Across Different Dataset Sizes and Models}
\label{tab:acc_results}
\begin{tabular}{lccccc}
\toprule
\textbf{Size} & \textbf{SMMT} & \textbf{3MT} & \textbf{ADDFformer} & \textbf{FusionNet} & \textbf{CNN} \\
             & \textbf{(Ours)} & {[8]} & {[20]} & {[21]} & {[22]} \\
\midrule
100\% & \textbf{97.05} & 90.28 & 88.20 & 94.28 & 80.24 \\
80\%  & \textbf{94.20} & 87.12 & 83.82 & 86.37 & 77.27 \\
60\%  & \textbf{91.28} & 86.25 & 77.58 & 79.25 & 70.57 \\
40\%  & \textbf{86.52} & 82.17 & 74.20 & 76.83 & 70.55 \\
20\%  & \textbf{84.96} & 78.92 & 71.92 & 75.15 & 68.53 \\
\bottomrule
\end{tabular}
\end{table}

To further analyze model performance, we examine the confusion matrix under the full-data (100\%) setting. As shown in Fig. 4, SMMT correctly classifies the vast majority of AD and CN cases, resulting in an overall accuracy of 97\%. The classification report indicates high precision (0.98 for CN, 0.95 for AD), recall (0.98 for CN, 0.95 for AD), and F1-score (0.98 for CN, 0.95 for AD), reflecting strong and balanced performance across both classes.

\begin{figure}[htbp]
    \centering
    \includegraphics[width=\linewidth]{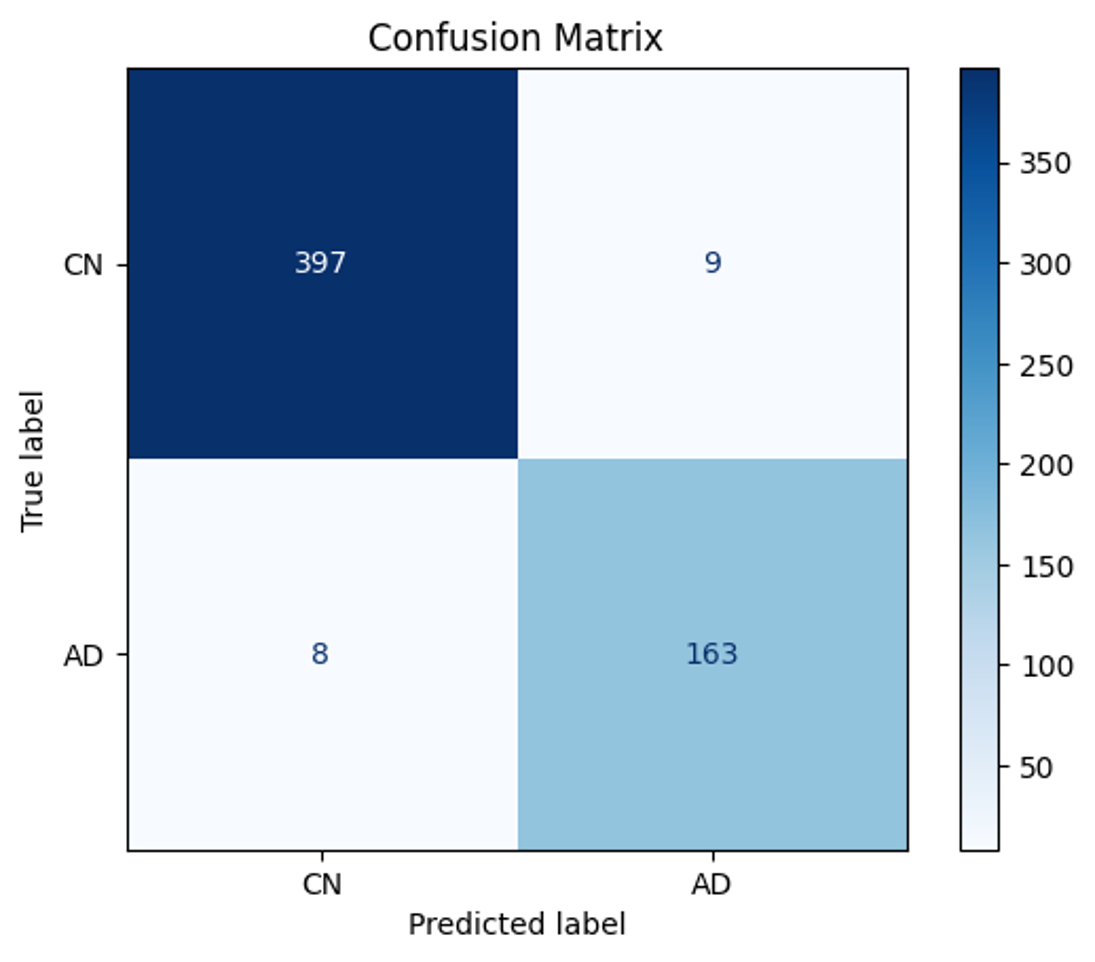} 
    \caption{Confusion matrix for SMMT under 100\% dataset size.}
    \label{fig:Confusion matrix}
\end{figure}
\FloatBarrier

Following the confusion matrix analysis, we further evaluate clinically important diagnostic metrics, including Sensitivity, Specificity, and Area Under the ROC Curve (AUC). These indicators are especially relevant in real-world healthcare applications, where minimizing false negatives (missed AD diagnoses) is critical.

\begin{table}[htbp]
\centering
\caption{Diagnostic Metrics Under 100\% Dataset Setting}
\label{tab:diagnostic_metrics}
\resizebox{\linewidth}{!}{%
\begin{tabular}{lccc}
\toprule
\textbf{Model} & \textbf{Sensitivity (\%)} & \textbf{Specificity (\%)} & \textbf{AUC} \\
\midrule
CNN-only [22]               & 78.45  & 80.22  & 0.852 \\
FusionNet [21]              & 91.01  & 93.24  & 0.956 \\
ADDFformer [20]             & 91.87  & 91.53  & 0.948 \\
3MT (Baseline) [8]          & 93.64  & 93.81  & 0.965 \\
\textbf{SMMT (Ours)}                   & \textbf{96.31} & \textbf{97.58} & \textbf{0.986} \\
\bottomrule
\end{tabular}%
}
\end{table}

As shown in Table~\ref{tab:diagnostic_metrics}, SMMT achieves the highest sensitivity, specificity, and AUC among all models. These results confirm that SMMT not only improves accuracy and robustness but also ensures balanced and clinically meaningful predictions for both AD and CN populations.

\begin{figure}[htbp]
    \centering
    \includegraphics[width=\linewidth]{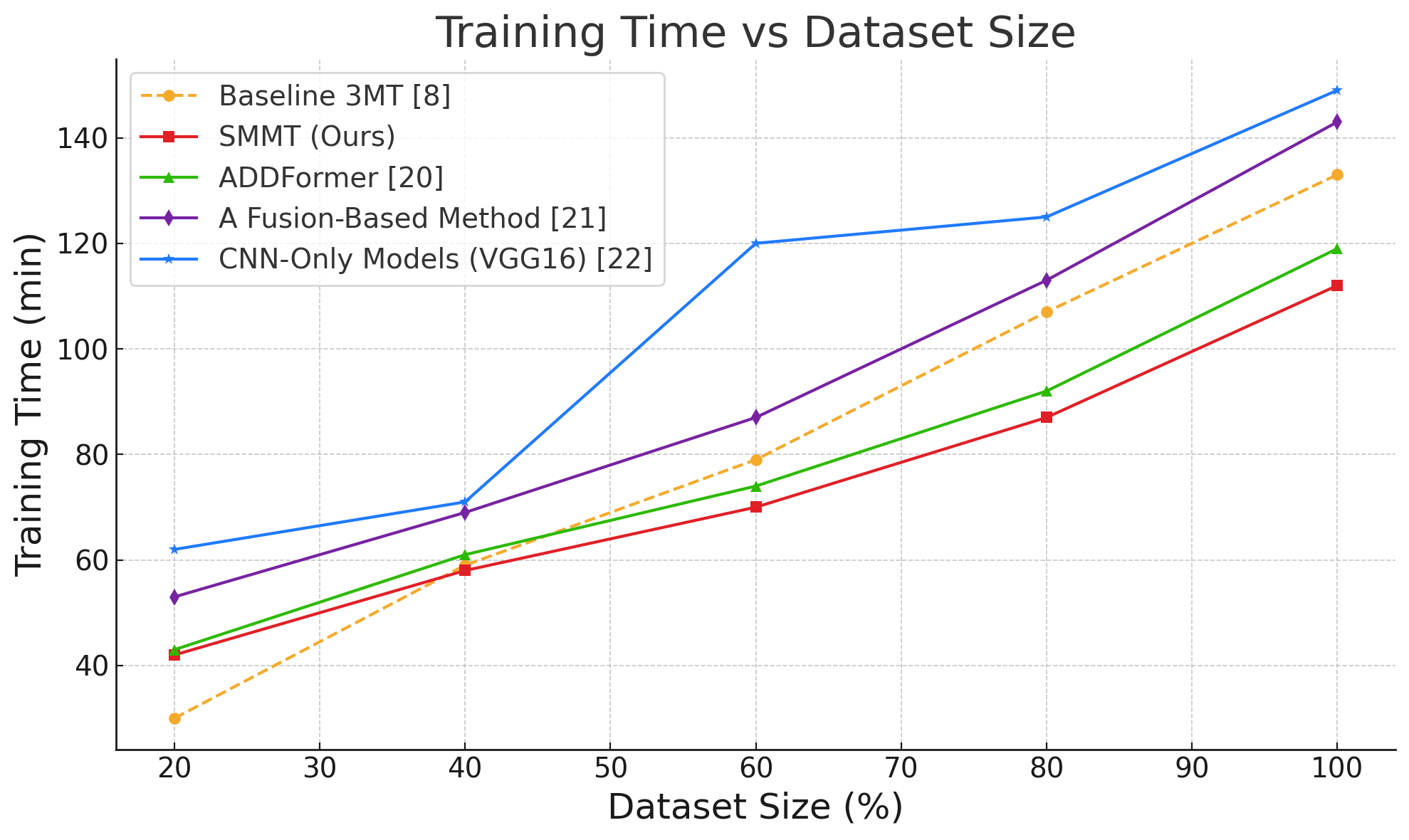} 
    \caption{Training time (minutes) vs. dataset size.}
    \label{fig:training_time}
\end{figure}

We also evaluate the computational efficiency of each model by comparing training time across varying dataset sizes. As shown in Fig.~\ref{fig:training_time}, the CNN-only model (VGG16) consistently exhibits the longest training time. While VGG16 has a straightforward sequential structure, its deep convolutional layers and large parameter count result in substantial computational cost, particularly compared to more efficient attention-based models like SMMT. In contrast, our SMMT model achieves the lowest training time among all Transformer-based and fusion methods, validating the effectiveness of sparse attention in reducing computational overhead. Compared to the baseline 3MT, SMMT demonstrates better scalability as dataset size increases.

Overall, the results validate that SMMT offers the best trade-off between classification performance and training efficiency, making it particularly suitable for deployment in real-world clinical settings where both robustness and computational cost are important considerations.

\subsection{Energy Efficiency Comparison}
In addition to classification accuracy and training time, we further evaluate the energy efficiency of each model by analyzing its power consumption across major hardware components—CPU, GPU, and RAM. This analysis is performed using the \texttt{CodeCarbon} framework, which monitors real-time energy usage during training and aggregates component-wise consumption over time [24].

Figure 6 presents the average energy consumption per epoch for both the baseline 3MT and our proposed SMMT model. It is evident that SMMT maintains a lower and more stable energy profile throughout the training process. In contrast, 3MT demonstrates a higher and more fluctuating consumption pattern, especially in the early and final stages of training.

\begin{figure}[htbp]
    \centering
    \includegraphics[width=\linewidth]{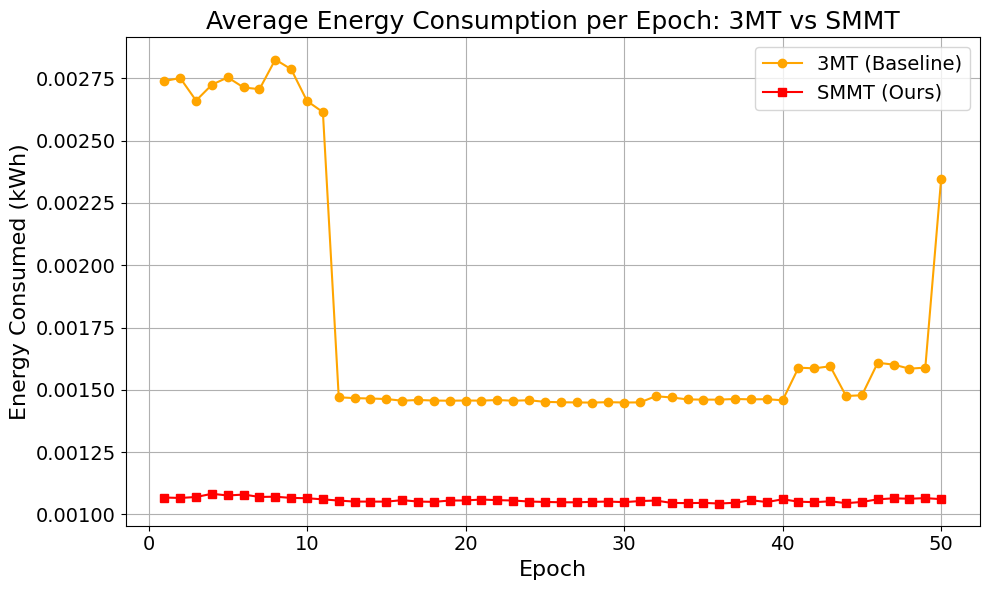} 
    \caption{Average energy consumption per epoch for 3MT and SMMT across training.}
    \label{fig:energy_curve}
\end{figure}

Compared to the 3MT baseline, which exhibits energy fluctuations particularly at the beginning and end of training, our SMMT model demonstrates a significantly more stable energy profile across all epochs. This stability arises from the integration of sparse attention, which reduces unnecessary computation by restricting attention to token clusters, and the use of masking, which regularizes training dynamics. These techniques not only reduce total energy usage but also result in smoother GPU and CPU utilization over time, leading to more predictable and sustainable energy demand throughout the training process.

To provide a more quantitative breakdown, we further summarize the total energy usage over 250 training epochs (5-fold cross-validation $\times$ 50 epochs) in Table~\ref{tab:energy_summary}. The results clearly indicate that SMMT is more energy-efficient across all hardware components. Specifically, SMMT reduces GPU energy consumption from 0.283977~kWh to 0.159642~kWh ($-43.8\%$), CPU usage from 0.108489~kWh to 0.071179~kWh ($-34.4\%$), and RAM usage from 0.051035~kWh to 0.033485~kWh ($-34.4\%$). Overall, SMMT achieves a 40.4\% reduction in total energy consumption compared to the 3MT baseline.

\begin{table}[htbp]
\centering
\caption{Component-Wise Energy Consumption}
\label{tab:energy_summary}
\resizebox{\linewidth}{!}{%
\begin{tabular}{lccc}
\toprule
\textbf{Component} & \textbf{3MT (Baseline)} & \textbf{SMMT (Ours)} & \textbf{Reduction (\%)} \\
\midrule
CPU  & 0.108489~kWh  & 0.071179~kWh  & 34.4\% \\
GPU  & 0.283977~kWh  & 0.159642~kWh  & 43.8\% \\
RAM  & 0.051035~kWh  & 0.033485~kWh  & 34.4\% \\
\midrule
\textbf{Total} & \textbf{0.443501~kWh} & \textbf{0.264306~kWh} & \textbf{40.4\%} \\
\bottomrule
\end{tabular}%
}
\end{table}

In addition to energy usage, we estimate the total $CO_2$ emissions generated during training. Based on the carbon intensity of Taiwan's power grid (0.502~$kgCO_2$/$kWh$), the 3MT model emits approximately 0.2226~$kgCO_2$, while our SMMT model emits only 0.1327~$kgCO_2$, resulting in a 40.3\% reduction in carbon footprint. Although the absolute difference of 0.0899~$kgCO_2$ may seem modest, it is roughly equivalent to charging a smartphone 18 times or keeping a 60-watt lightbulb on for over 30 hours. This emphasizes SMMT’s advantage not only in predictive performance and computational efficiency, but also in long-term environmental sustainability, especially when scaled to large datasets or frequent model retraining scenarios.

These results confirm that the proposed SMMT architecture not only enhances diagnostic performance in low-resource medical settings but also offers substantial reductions in environmental impact and computational cost. This makes SMMT particularly suitable for sustainable AI deployments in real-world healthcare environments.

\subsection{Ablation Study}

To assess the contribution of each architectural component in our proposed SMMT model, we perform ablation studies by selectively disabling Sparse Attention and Masking modules. The baseline model 3MT [8] is used as a reference. We compare the full version of SMMT with two reduced variants: one without sparse attention (that is, reverting to dense self-attention), and another without masking (that is, training on full data).

Table~\ref{tab:ablation} summarizes the accuracy and training time of each configuration across five levels of dataset availability (from 20\% to 100\%). When sparse attention is removed, the accuracy decreases slightly, while training time increases significantly, particularly under full-data conditions (from 112 minutes to 147 minutes at 100\% data). This validates the computational efficiency gained by sparse attention. On the other hand, removing the masking mechanism results in a slight reduction in training time, but leads to noticeable drops in accuracy, especially under low-data conditions. For instance, at 20\% data, accuracy drops from 84.96\% to 80.85\% without masking. This demonstrates that the masking mechanism plays an important role in improving model robustness and generalization under data-limited settings.

While the primary motivation for introducing sparse attention was to reduce computational complexity, our ablation results indicate that it may also contribute to improved generalization performance. Specifically, when the sparse attention module is removed, the model exhibits a noticeable decrease in classification accuracy despite having access to full token-to-token interactions. This observation suggests that the sparsity constraint may implicitly serve as a regularization mechanism by limiting unnecessary or noisy attention connections. Such behavior encourages the model to prioritize salient and semantically meaningful token relationships, thereby reducing overfitting. This implicit regularization effect is conceptually aligned with techniques such as dropout and attention masking, particularly beneficial in settings with limited or heterogeneous data.

\begin{table}[htbp]
\centering
\caption{Ablation Study: Accuracy (\%) / Training Time (min)}
\label{tab:ablation}
\begin{tabular}{lcccc}
\toprule
Size & 3MT & w/o Sparse & w/o Masking & SMMT (Full) \\
\midrule
100\% & 90.28 / 133 & 91.35 / 147 & 95.42 / 108 & \textbf{97.05 / 112} \\
80\%  & 87.12 / 106 & 89.14 / 140 & 93.18 / 91  & \textbf{94.20 / 107} \\
60\%  & 86.25 / 79  & 88.37 / 96  & 91.62 / 68  & \textbf{91.28 / 87} \\
40\%  & 82.17 / 59  & 85.03 / 78  & 84.43 / 56  & \textbf{86.52 / 70} \\
20\%  & 78.92 / 41  & 84.85 / 51  & 80.85 / 37  & \textbf{84.96 / 45} \\
\bottomrule
\end{tabular}
\end{table}

Overall, the results confirm that both sparse attention and masking contribute independently to the performance of SMMT. Sparse attention improves computational efficiency with minimal impact on accuracy, while masking significantly enhances generalization, particularly when training data is limited. The combination of both allows SMMT to achieve a favorable balance between efficiency and effectiveness.

\begin{figure}[htbp]
    \centering
    \includegraphics[width=\linewidth]{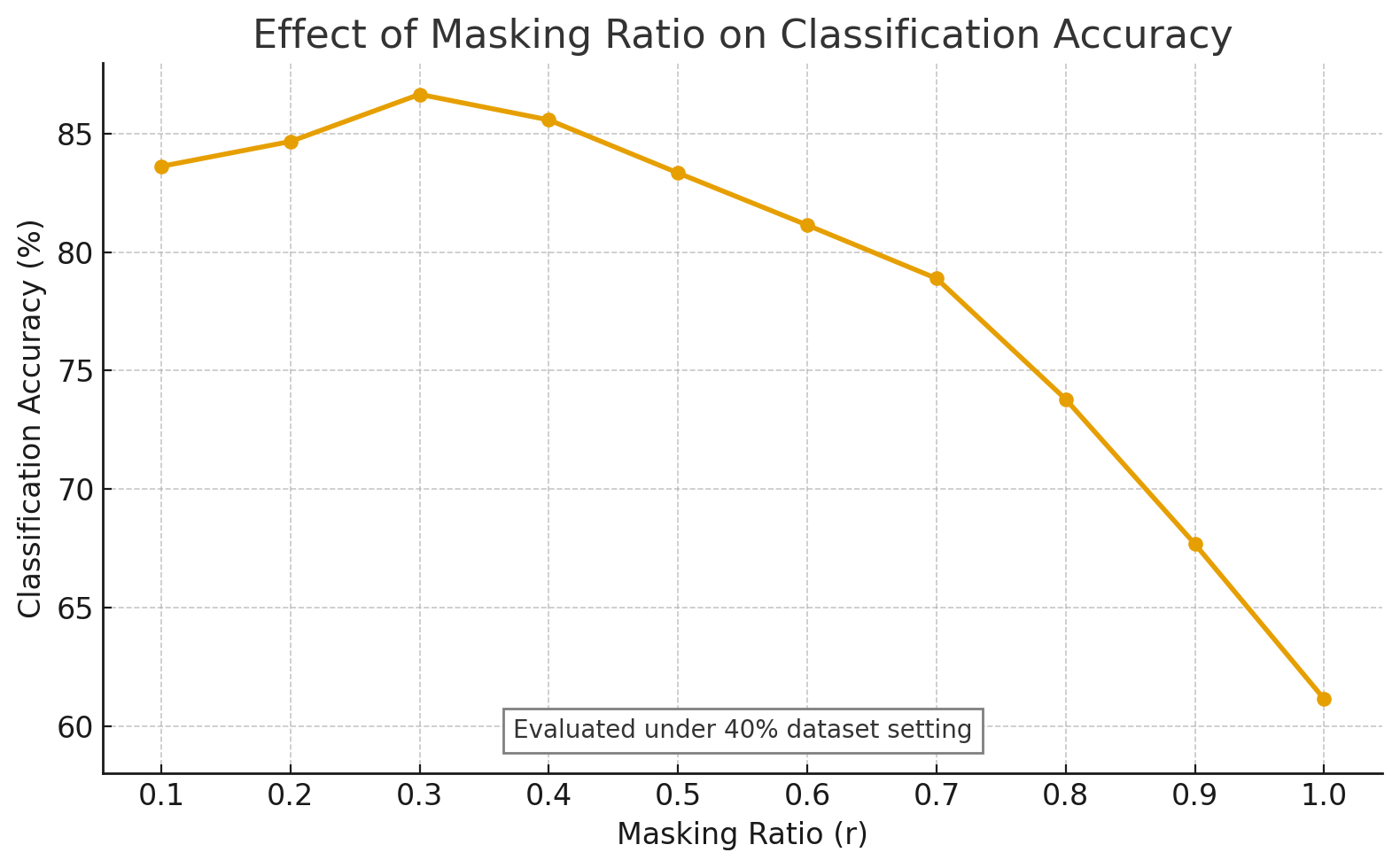} 
    \caption{Effect of masking ratio \(r\) on classification accuracy under the 40\% dataset setting. Peak accuracy is observed at \(r=0.3\), with performance degrading as masking becomes too aggressive.}
    \label{fig:masking_ablation}
\end{figure}

To further justify the choice of the masking ratio \(r=0.3\), we conduct an ablation study varying \(r\) from 0.1 to 1.0 under the 40\% dataset setting. As shown in Fig.~\ref{fig:masking_ablation}, model performance peaks at \(r=0.3\), and begins to decline as the masking becomes more aggressive. When the masking ratio exceeds 0.6, accuracy drops significantly and approaches the majority class baseline (60\%), indicating that excessive masking removes critical information and limits the model’s ability to generalize. This empirical result supports the selection of a moderate masking level that balances regularization and information retention.

\section{Future Work and Conclusion}
This study focuses solely on binary classification between AD and CN individuals. While including Mild Cognitive Impairment (MCI) cases would provide a more comprehensive clinical perspective, we deliberately exclude them in this work due to two practical constraints. First, MCI samples constitute a relatively small portion of the dataset, leading to class imbalance and higher risk of overfitting, especially under limited-data settings. Second, introducing a third class increases task complexity and requires additional computational resources and hyperparameter tuning, which may detract from our focus on architectural benchmarking and energy-efficiency analysis. We acknowledge the clinical importance of MCI and plan to extend our framework to include MCI and progressive staging in future work.

In this work, we proposed SMMT, a Sparse Multi-Modal Transformer with Modality-Wise Masking, designed for accurate and efficient Alzheimer’s Disease classification under limited data and resource constraints. By introducing cluster-based sparse attention, we reduced the time complexity of intra-modal processing while maintaining strong representational capacity. The proposed masking strategy further enhanced model robustness by simulating missing modalities during training, improving generalization in real-world clinical settings.

Comprehensive experiments on the ADNI dataset demonstrated that SMMT outperforms multiple state-of-the-art baselines including 3MT, ADDFformer, and CNN-only models not only in classification accuracy but also in energy efficiency. Notably, SMMT achieved 97.05\% accuracy on full data and 84.96\% using only 20\% of the training set, while reducing energy consumption by over 40\% compared to 3MT.

In future work, we plan to extend our framework to support longitudinal modeling for AD progression prediction and explore adaptive modality selection mechanisms to dynamically handle variable modality availability at inference time. Additionally, we aim to validate SMMT on larger and more diverse clinical datasets to further assess its generalizability across patient populations and healthcare systems.

\ifCLASSOPTIONcaptionsoff
  \newpage
\fi

\end{document}